# Assessing cloudiness in nonwovens

M. Godehardt[1], A. Moghiseh[1], C. Oetjen[1,2], J. Ohser[3], S. Ringger[4], K. Schladitz[1], I. Windschiegel[4]

[1]Fraunhofer-Institut für Techno- und Wirtschaftsmathematik, Kaiserslautern, Abteilung Bildverarbeitung
[2]Universität Kaiserslautern, Fachbereich Mathematik
[3]Hochschule Darmstadt, Fachbereich Mathematik und Naturwissenschaften
[4]Deutsche Institute für Textil- und Faserforschung Denkendorf, Kompetenzzentrum Chemiefasern und Vliesstoffe

## Abstract:

The homogeneity of filter media is important for material selection and quality control, along with the specific weight (nominal grammage) and the distribution of the local weight. Cloudiness or formation is a concept used to describe deviations from homogeneity in filter media. There are various image analysis methods for measuring cloudiness differing in the exact definition of this term. Cloudiness concepts based on the range of interaction [1], the coefficients of the Laplace pyramid [2,3], or the power spectrum [4,5] have been suggested. Measured cloudiness is reproducible. However, cloudiness measured with varying instruments usually cannot be compared. We have developed a theoretically sound cloudiness index and a method to measure it from the power spectrum. The eligible frequency band depends on the image acquisition. Reproducibility of the method is guaranteed as long as the nonwoven sample attenuates the light proportionally to the material thickness. Our method is hence well suited to build a technical standard on it.

Following [4,5], we suggest to derive the cloudiness index from the power spectrum of the relative local areal weight, integrated over a selected frequency range. More precisely, we denote by areal weight the mean weight per unit area, measured in $g/m^2$. The local areal weight is the weight in an infinitesimally small area element, the local weight as a function of the planar position, measured in $g/m^2$, too. The relative local areal weight is then the fraction of these two - local areal weight divided by areal weight (also called relative or normalized local grammage) and unit-less. Cloudiness is then quantified as the integral of the relative local areal weight over the chosen frequency range. This cloudiness measure has various advantages over popular alternatives, both in terms of the information contained and the robustness of the measurement [5]. The power spectrum captures the energy density in a broad spectral range. Moreover, under certain conditions, the structure of a nonwoven is fully characterized by the areal weight, the variance of the local areal weight, and the power spectrum. Consequently, the power spectrum is the parameter that exclusively reflects the cloudiness.

Here, we address questions arising from practical application. The most prominent is the choice of the spectral band to properly capture a nonwoven's cloudiness relevant for the physical properties as well as meeting visual perception. The band certainly depends on the characteristic "size of the clouds", but is limited by the chosen imaging method. Choosing coefficients of the Laplace pyramid instead, does not simplify the task. Moreover, the values of neighbouring coefficients do depend on each other and thus contain redundant information.

The spectral band of the measured power spectrum is limited due to the limited size and lateral resolution of images. The theoretical lower bound of the spectral band is the inverse half diagonal length of the image. However, measurement errors are very large for frequencies just above this limit. The theoretical upper limit for the frequencies is the reciprocal of twice the pixel size - the smallest wavelength for which the energy density can be determined. In practice, the smallest wavelength should be a multiple of the pixel size.

To summarize, the cloudiness index based on the power spectrum of the relative local areal weight is theoretically well founded and can be robustly measured from image data. Choosing the spectral band allows to capture the cloudiness either visually perceived or found to be decisive for product properties. It is thus well suited to build a technical standard on it.

# Keywords:
image analysis, homogeneity, power spectrum

# 1. Method

## Cloudiness based on the power spectrum

The macroscopic properties of nonwovens are decisively determined by the mean fiber diameter, the fiber diameter distribution, the fiber curvature, the mean weight of the material per unit area and the material thickness. Deviation from homogeneity (spatial fluctuation or inhomogeneity) is another morphological feature with significant influence on macroscopic technological properties like the permeability for liquids or gases. The less homogeneous a nonwoven is (while all other morphological features are kept constant), the more inhomogeneous gets the flow through it. Thus, the macroscopic permeability can increase significantly.

Nevertheless, it is far from obvious, what exactly is meant by this (in)homogeneity. Is there a well-defined characteristic quantifying it? If yes, can this characteristic be measured reproducibly? Physically or using image analysis? In fact, inhomogeneity of a random structure like a nonwoven is a concept that is difficult to pin down. Thus, a variety of characteristics have been suggested to describe the inhomogeneity or cloudiness of nonwovens.

First, there is the scattering intensity of coherent X-rays [6] or β-radiation [7] that can be measured physically based on small-angle scattering. All others are characteristics that can be estimated image analytically based on optical microscopy. The prerequisite for all of them is the nonwoven being sufficiently thin for the light being attenuated following Beer-Lambert's law. That is, the transmitted light images capture the spatial distribution of the local weight. The autocorrelation function and the power spectrum [4,5,8] estimated based on transmission microscopy images are equivalent to the scattering intensity [5]. More precisely, these quantities can be transformed into each other and thus contain the same information about the inhomogeneity. In [5] it was also shown, that the variance of the integrated local areal weight in a square or circular window [6] and the range of interaction [1] can be derived from the scattering intensity, the autocorrelation function or, in particular, the power spectrum.

Finally, there are methods collecting structural information from several scales: The Laplace pyramid [2,3,9] uses Gaussian filters of varying width, the lacunarity [10] square subimages (boxes) of varying edge lengths. This approach seems appealing at first glance. However, as shown in [5], the coefficients of the Laplace pyramid, and thus the information about the cloudiness from the considered scales are not independent. The lacunarity shares this undesirable property. The Laplace pyramid can be derived from the power spectrum [5]. The lacunarity however is essentially the root mean square contrast in the square boxes. For anisotropic nonwovens (i.e. nonwovens whose structure is not on average rotation invariant), the lacunarity thus depends on the orientation of these square boxes. For increasing edge length of the boxes, the lacunarity converges to the coefficient of variation of the local areal weight which is not the inhomogeneity on the mesoscale. We therefore rule out lacunarity for characterizing cloudiness.

To summarize, all well-defined characteristics can be traced back to the power spectrum. We therefore focus on its image-analytical estimation in the following.

## Estimating the power spectrum from transmission microscopy images

The relative local areal weight of a nonwoven equals, up to a constant factor, the logarithm of the gray values in the transmission microscopy image if the light attenuation follows Lambert-Beer's law, reflection and scattering are negligible, and no pixel value is zero [11].
If these conditions are fulfilled, the power spectrum can be estimated reproducibly, provided the lateral resolution is in the range of the fiber diameter and the scanned total area of the nonwoven is large enough. If the fiber material absorbs light too strongly or the nonwoven is too thick for the light attenuation to follow Lambert-Beer's law, the power spectrum can still be estimated reproducibly. However, in that case, all image acquisition parameters like intensity and spectrum of the applied light, numerical aperture of the objective lens, optical magnification, and camera gain have to be kept constant. Finally, X-ray transmission imaging is an alternative, although for quality assurance of nonwovens probably too costly.

The power spectrum is calculated efficiently using a fast Fourier transform. In fact, it is the squared magnitude of the complex-valued pixels of the Fourier transformed image. When interpreting the power spectrum image analytically, the change of pixel size and pixel values due to the Fourier transform has to be taken into account. As a consequence, the pixel size in frequency space has unit m-1 and for the power spectrum m² [5]. Moreover, edge effects due to the Fourier transform periodically extending the image have to be corrected with the help of the window function.

Averaging estimates of the power spectrum based on images of several fields of view of the same nonwoven, is an obvious strategy to reduce the statistical estimation error. Nonwovens are macroscopically homogeneous. Thus, alternatively, the power spectrum can be estimated as follows:  First calculate the pixel-wise mean of the images. Subsequently, estimate the power spectrum from this mean. In other words, calculating the mean and estimating the power spectrum can be interchanged. This accelerates the computation considerably.

## The cloudiness index

Usually, it is not the complete power spectrum that is of interest but the integral of the power spectrum over a specified frequency band - the cloudiness index (CLI). The integral of the power spectrum over the full frequency range is 1. Hence, the CLI is a number between 0 and 1 and can be expressed in percent. The higher the CLI, the less homogeneous is the investigated nonwoven in the considered frequency band.

Let $\omega(x)$ be the local areal weight in the point $x$ in the plane – a random function on the plane. Denote by $\overline{\omega}$ its expectation, that is the areal weight of the nonwoven, and by $\sigma_\omega$ its standard deviation. The correlation function $k(x)$ of $\omega(x)$ is the expected value of $(\omega(y) - \overline{\omega})(\omega(y+x) - \overline{\omega})/\sigma_\omega^2$ for any point $y$ in the plane. That is, $k(x)$ describes the joint deviation of the local areal weights expected to be observed at two points in the plane that are the vector $x$ away from each other. The power spectrum of the relative areal weight
$$(\omega(x) - \overline{\omega})/\sigma_\omega$$
is then just the Fourier transform
$$\hat{k}(\xi) = \frac{1}{2\pi} \int_{\mathbb{R}^2} k(x) e^{-ix\xi}\, dx$$
of the correlation function over the plane $\mathbb{R}^2$. Finally, we consider its rotation mean $\hat{k}_1(\|\xi\|) = \hat{k}(\xi)$. The integral of $\hat{k}_1(\rho)$ over the chosen frequency band finally yields our CLI. For more mathematical details we refer to [5]. Note that the power spectrum $\hat{k}_1(\rho)$ is a function of the frequency $\rho$ (measured in µm$^{-1}$) and has unit µm$^2$.

This definition of cloudiness clearly simplifies the comparison of structural inhomogeneity of nonwovens from varying sources. It induces however also the requirement to always specify the frequency band for the comparison. This frequency band certainly depends on how finely structured the investigated nonwovens are. The finer the nonwoven is, in particular the thinner the fibers are, the further the band should be shifted to higher frequencies. If the frequency band shall be adapted to meet the subjective visual perception of cloudiness, i.e. the perceived ``typical´´ cloud size, it helps to express the frequency band in terms of wavelengths instead. In industrial quality assurance, manufacturer and customer have to agree on the relevant frequency band. Finally, the choice of the frequency band can also be driven by technical specifications.

## Choice of lateral resolution and frequency band

The systematic error (bias) of an estimated power spectrum depends primarily on the lateral resolution of the microscope used, i.e. on the numerical aperture $n$ of the objective lens. In good microscopes, the pixel size is slightly smaller than the lateral resolution. To resolve the fibers and thus to limit the systematic error caused by the resolution, the pixel size should not be less than half of the fiber diameter.

On the other hand, the statistical error of the estimated power spectrum depends on the scanned total area of the nonwoven and the chosen frequency. Obviously, the larger the total area, the smaller the statistical error. The statistical error also decreases with increasing frequency $\rho$, i.e. with decreasing wave length $\gamma = 2\pi/\rho$ . Quantitatively expressed, this means: The statistical error is inversely proportional to the area of the field of view of the microscope, morphologically eroded by a line segment of length $\gamma$.

In summary, both, a high lateral resolution and a large field of view of the microscope are needed to limit the total error (bias plus statistical error) when estimating the cloudiness of nonwovens. Repeated measurements for several fields of view can nevertheless replace to some extent the latter.

Note that averaging the image-analytically determined power spectra over a large number of fields of view is not only needed but also quite natural as the comparison with small-angle X-ray scattering (SAXS) shows. As mentioned above, SAXS is the only method for physically measuring the power spectrum. SAXS measures the scattering intensity on thin layers of materials, which in principle corresponds to the power spectrum determined by image analysis. However, even when using SAXS, averaging over several sample positions is needed to control the statistical error. The latter has led to the development of scanning small-angle X-ray scattering (sSAXS), a method that is nowadays successfully used to characterize structural inhomogeneities of materials [12].

## 2. Cloudiness of simulated structures

The way cloudiness of nonwovens expresses itself in transmission microscopy images, can be modelled based on random Gaussian fields (GRF), generated by the so-called dilution or spectral methods. The latter builds on an appropriately chosen power spectrum. This also relates characterization and modelling of cloudiness: An estimated power spectrum can be input for generating a GRF by the spectral method.

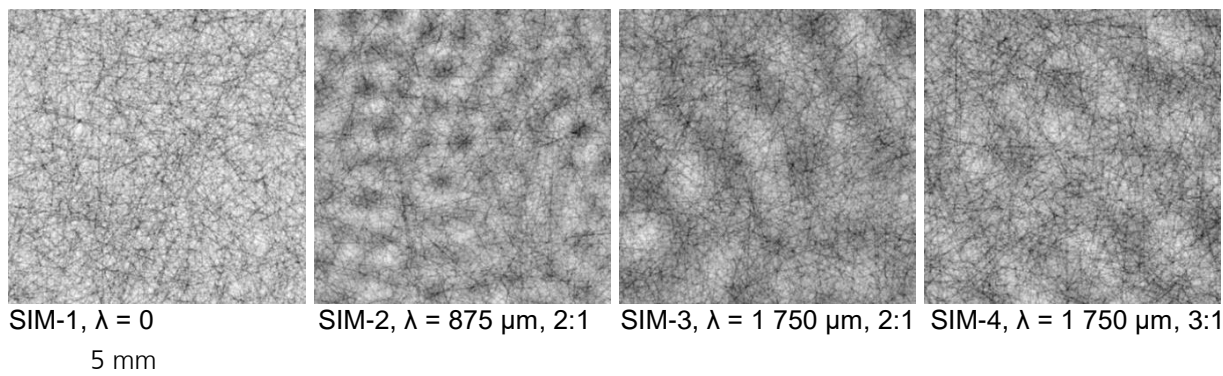

SIM-1, λ = 0    SIM-2, λ = 875 µm, 2:1    SIM-3, λ = 1 750 µm, 2:1    SIM-4, λ = 1 750 µm, 3:1

5 mm

**Figure 1:** Simulated transmission microscopy images of nonwovens with varying cloudiness, 1 024 x 1 024 pixels of size 7 µm. Modeling based on superposing a GRF generated by dilution of segments of diameter 42 µm and exponentially distributed length with mean 896 µm and a GRF generated by the spectral method with a Bessel autocorrelation function.

Figure 1 yields simulated nonwoven images with varied cloudiness. The images are generated by dilution with segments of diameter 42 µm and exponentially distributed random lengths with mean 896 µm. In this model, the segments model filaments. The cloudiness is altered by superposing the simulated nonwoven by GRFs generated by the spectral method and using Bessel autocorrelation functions with parameter λ = 0 for SIM-1, λ = 875 µm for SIM-2, and λ = 1 750 µm for SIM-3 and SIM-4. Clearly, λ = 0 means superposition with a constant. Thus, SIM-1 is free of additional cloudiness. SIM-2 and SIM-3 are superpositions of the fiber model with the GRF in the ratio 2:1, while a ratio of 3:1 was chosen for SIM-4.

The CLI derived for the frequency range from 0.002 µm$^{-1}$ to 0.010 µm$^{-1}$ are 10.15%, 42.38%, 32.28%, and 24.95%. As expected, the CLI increases from SIM-1 to SIM-2. The CLI of SIM-3 decreases again and even further for SIM-4. The CLI of SIM-4 could be expected to be lower than the one of SIM-3, too, due to the lower weighting of the GRF. Subjectively perceived cloudiness is therefore not only a question of scale as represented in our example by the parameter λ but also of the intensity represented by the weighting of the underlying GFR. Direct comparison of the simulated nonwovens SIM-2 and SIM-3 emphasizes however also, how crucial the CLI depends on the chosen frequency band. Choosing just 0.002 µm$^{-1}$ to 0.006 µm$^{-1}$ results in CLI 42.06%, 4.75%, 27.27%, and 18.67% for SIM-1 to SIM-4. This changes the ranking from SIM-2 being cloudier than SIM-3 being cloudier than SIM-4 to CLI(SIM-3)>CLI(SIM-4)>CLI(SIM-2).

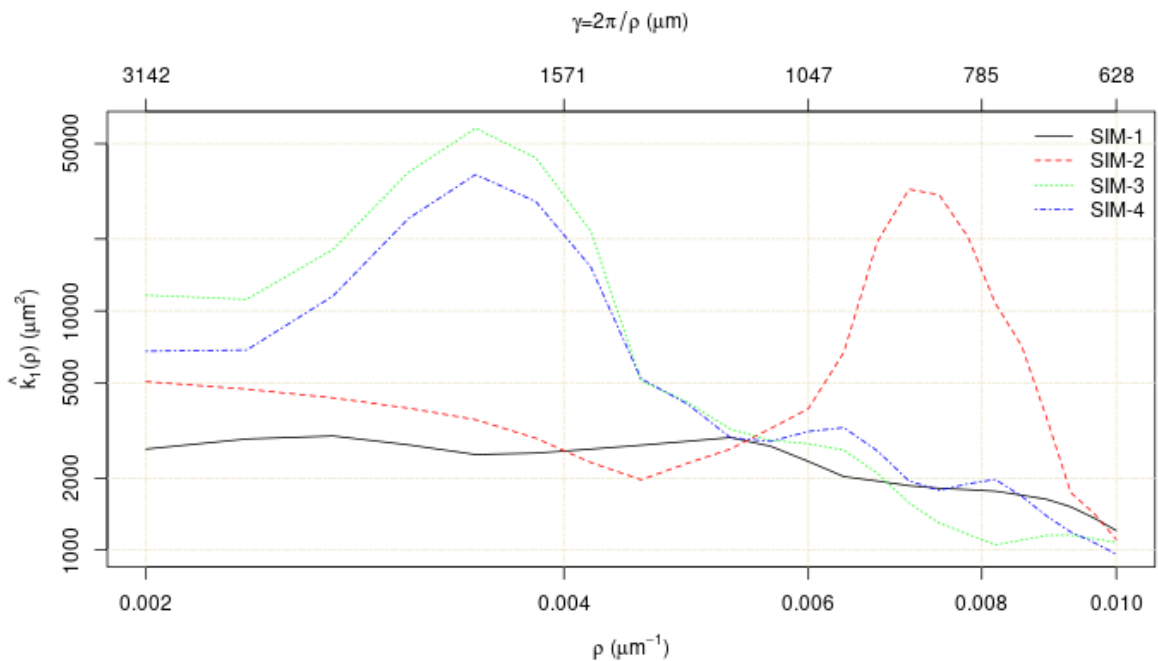

**Figure 2:** Rotation means of the power spectra for the simulated images SIM-1 to SIM-4 from Figure 1.

## 3. Cloudiness of real nonwoven samples

### Imaging

All nonwoven samples are imaged in transmission mode using a Leica stereo light microscope MZ16 equipped with a planapochromatic objective with numerical aperture 0.14 and working distance 55 mm. A Phlox-LedRGB-BL-100 x 100-S-Q-IR-24V, white, was used as light source. Exposure time was 2 ms. Images were recorded by a Basler acA4112-8gm detector with 4 096 x 3 000 pixels; sensor IMX304 CMOS, GigE, mono.

## Materials

Fiber deposition in nonwoven technologies is a chaotic process. Homogeneity therefore is always an issue for nonwoven products, especially crucial for example in filtration or separation applications. We apply the CLI on microscopic images of two groups of meltblown samples. One group made out of the go-to polymer polypropylene and the other made of high performance polyether ether ketone.

Polypropylene (PP) is one of the most commonly used polymers to be processed to nonwovens. Due to its low price and wide range of processability, it is ideal for mass production and for experimental setups alike. Most recently, PP meltblown nonwovens have been very much in the public eye, due to their application in pandemic face masks as functional filtration layer. In more advanced applications, for example blood filtration, product size is very small, which emphasizes the importance of homogeneity even more and requires new methods of evaluation.

The second group of samples is made of polyether ether ketone (PEEK) and represent high performance polymers. Due to its high thermal stability and its chemical resistance, the rather expensive PEEK is preferably used in challenging environments. Prominent examples are hot gas filtration or separator membranes in batteries and fuel cells. In all of these applications, thin layers with high homogeneity are desirable, if not of paramount importance.

Characteristics of the PP and the PEEK samples are summarized in Table 1 below.

**Table 1:** Sample characteristics summarized. Areal weight, median fiber diameter, macroscopic air permeability, and material thickness are measured based on the commonly used nonwoven standards.

| ID | Type | Areal weight | Median fiber diameter | Air permeability @200Pa | Material thickness | CLI |
|---|---|---|---|---|---|---|
| | | [g/m²] | [µm] | [l/m²/s] | [µm] | [%] |
| PP-1 | PP Meltblown | 70 | 2.9 | 870 | 800 | 42.8 |
| PP-2 | PP Meltblown | 40 | 2.1 | 740 | 600 | 41.2 |
| PP-3 | PP Meltblown | 40 | 2.0 | 770 | 700 | 41.3 |
| PP-4 | PP Meltblown | 50 | 2.5 | 700 | 800 | 49.0 |
| PEEK-1 | PEEK Meltblown | 10 | 1.5 | 4 800 | n.d. | 69.1 |
| PEEK-2 | PEEK Meltblown | 60 | 3.5 | 1 400 | 330 | 59.2 |
| PEEK-3 | PEEK Meltblown | 10 | 3.0 | 5 500 | 170 | 71.4 |
| PEEK-4 | PEEK Meltblown | 30 | 1.9 | 1 300 | 150 | 65.9 |

For the samples listed in Table 1, the CLI is estimated for the frequency range from 0.02 µm$^{-1}$ to 0.10 µm$^{-1}$ corresponding to wave lengths from 62.8 µm to 314 µm.

This choice is a compromise between the desired resolution of about half of the fiber diameter and the need to capture a sufficiently large field of view to allow for stable estimation. For fiber diameter 2.0 µm, pixel size 1.0 µm is thus desired. However, this induces fields of view about 50 times smaller than those obtained at pixel size 7.2 µm. Consequently, we would need about 500 instead of 10 fields of view to reach the same estimation precision for the CLI. Thus, the theory-based demand for resolving the fibers has been ignored in favor of statistical representativity. In any case, the images do reflect very well the visually perceived cloudiness of the nonwoven samples.

## Power spectra of the PP meltblown samples

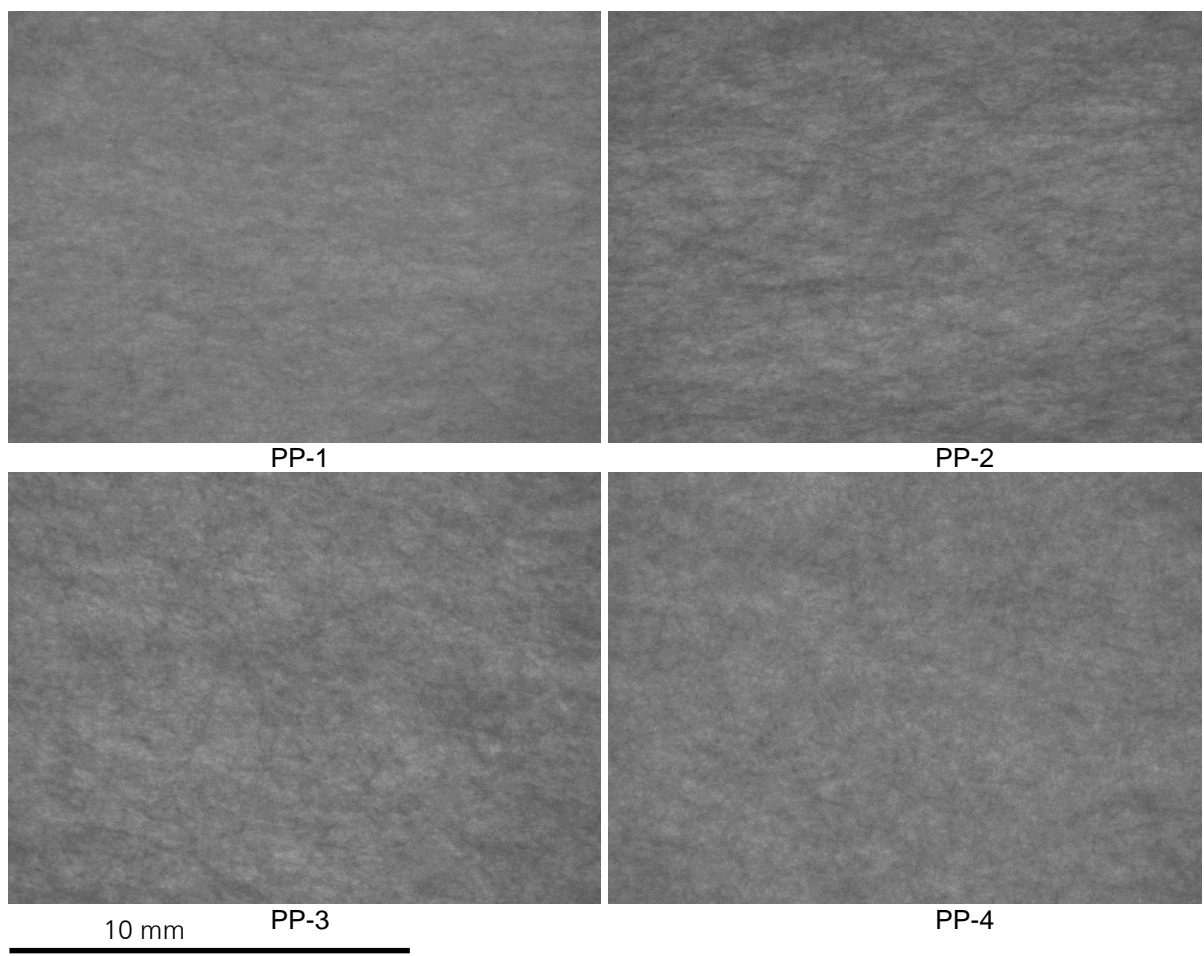

**Figure 3:** Transmission microscopy images of PP meltblown samples PP-1 to PP-4 with pixel size 7.2 µm. Pixel-wise averages over 10 images each. Rectangular fields of view of size 2 048 x 1 500 pixels cut from the original microscopic images.

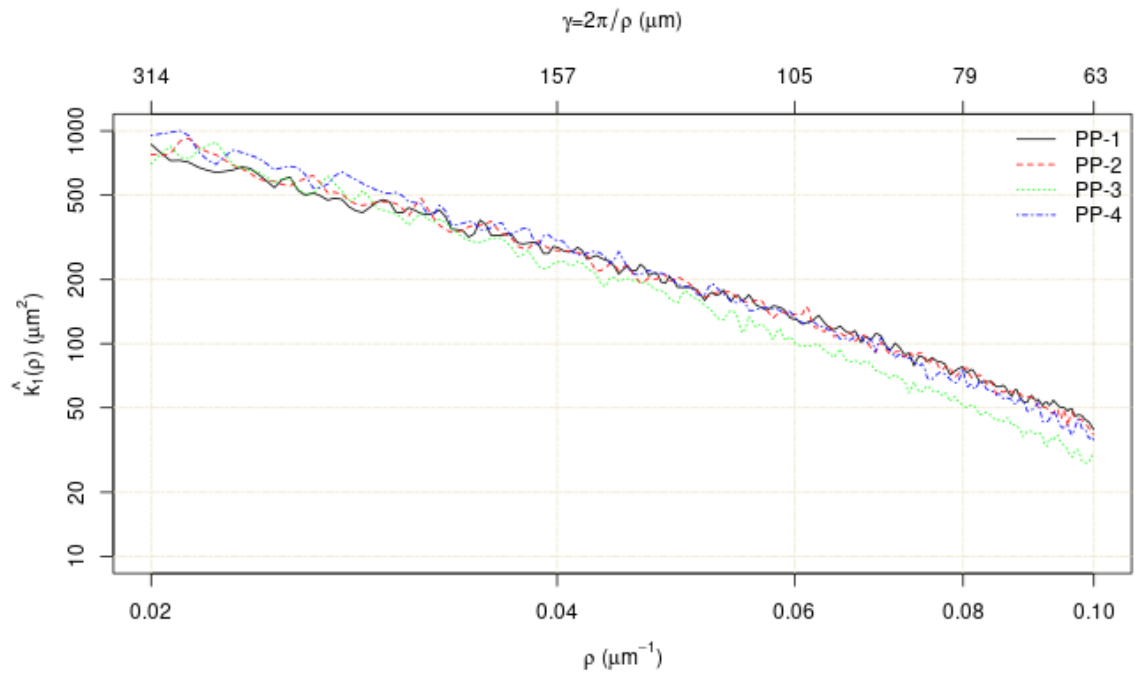

**Figure 4:** Rotation averages of the power spectra of the relative areal weight for the PP samples. Estimated based on the averaged transmission microscopy images from **Figure 3.**

## Power spectra of the PEEK meltblown samples

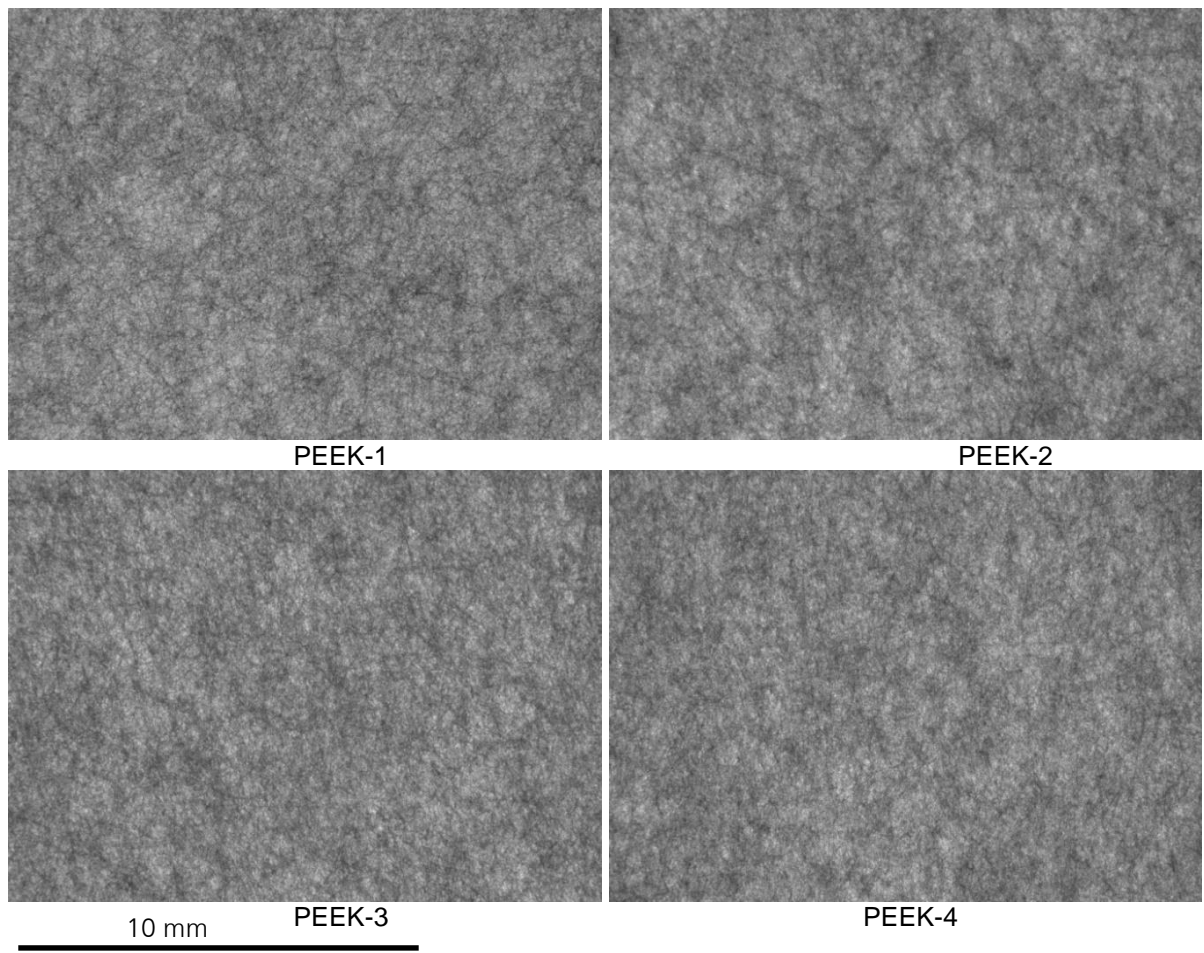

**Figure 5:** Transmission microscopy images of PEEK meltblown samples PEEK-1 to PEEK-4 with pixel size 7.2 µm. Pixel-wise averages over 10 images each. Rectangular fields of view of size 2 048 x 1 500 pixels cut from the original microscopic images.

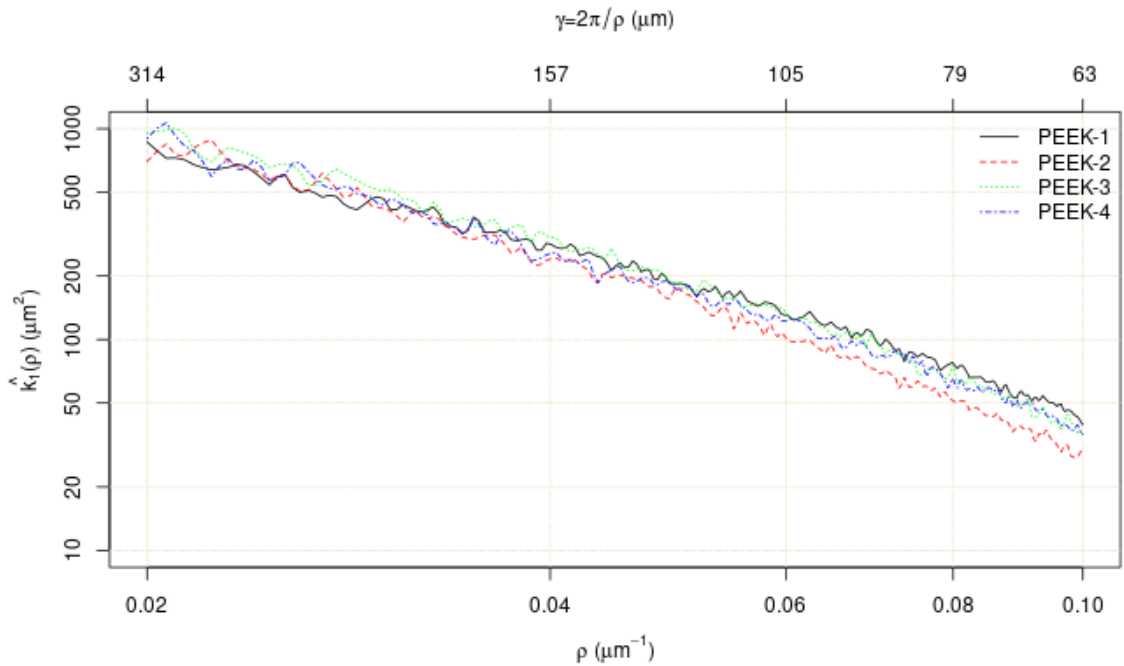

**Figure 6:** Rotation averages of the power spectra of the relative areal weight for the PEEK samples. Estimated based on the averaged transmission microscopy images from **Figure 5**.

## Estimated CLI

Based on the areal weights and the median fiber diameters given in Table 1, PEEK-3 is expected to be the most inhomogeneous sample due to its low areal weight while featuring rather thick fibers. In general, thinner fibers and higher areal weight are expected to increase the homogeneity. Thus, overall, the PEEK meltblown samples should be less homogeneous than the PP meltblown samples.

The CLI reported in the last column of Table 1 correspond to these expectations: The PEEK samples are cloudier than the PP ones and PEEK-3 is indeed the sample that yields the highest CLI (71.4 %). See also Figure 7 below.

The differences in cloudiness between the PEEK samples as well as between the PP samples are however rather small. In order to get more pronounced results, a significantly higher number of images have to be analyzed.

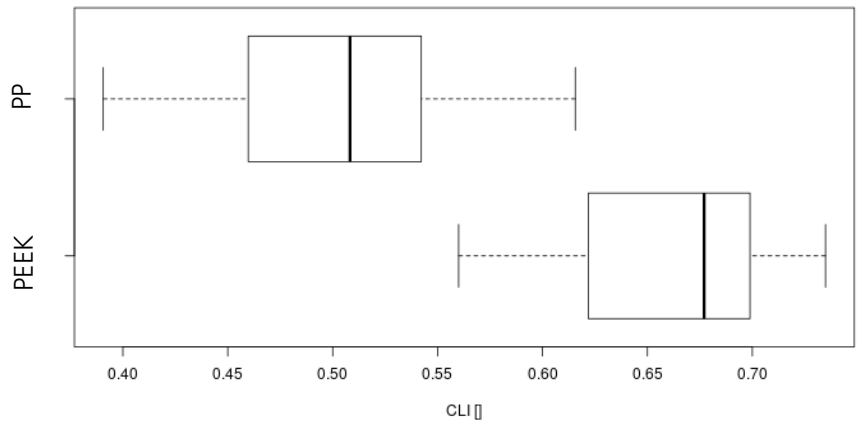

**Figure 7:** Box plot of the CLI measured based on 10 individual images each of the PP and the PEEK meltblown samples. The PEEK samples are on the chosen frequency range significantly cloudier than the PP samples.

# 4.  Conclusion

We suggest to measure inhomogeneity or cloudiness of nonwovens by a cloudiness index derived from the power spectrum of of the relative local areal weight. This choice is well motivated and mathematically sound [5]. Our CLI is unambiguous and objective.

Application to PP and PEEK nonwoven samples proves that the CLI yields the expected results. Simulated examples emphasize however also, that the choice of the frequency band is crucial for the resulting CLI.

The frequency band from 0. 02 µm$^{-1}$ to 0.10 µm$^{-1}$ first chosen purely empirically turned out to be practicable and to yield meaningful analysis results. Corresponding to wave lengths from approximately 60 µm to 300 µm it captures however rather the fine clouds. For a CLI capturing macroscopic thin places, coarser resolved images would be needed.

**Acknowledgement**: This research was supported by the project ``Product and quality optimization for protective clothing against infections made from nonwovens´´ (ProQuIV) within the Fraunhofer Society's anti-Corona programme. We thank Kai Taeubner for the microscopic imaging.

# References

[1] K. Zeng, J. G. Korvink (2000). The range of interaction for the characterization of cloudiness of nonwovens. WCC 2000 - ICSP 2000. 2000 5th International Conference on Signal Processing Proceedings. 16th World Computer Congress 2000, 2:1193-1200, doi: 10.1109/ICOSP.2000.891762

[2] J. Weickert (1999). A real-time algorithm for assessing inhomogeneities in fabrics, Real-Time Imaging 5:15-22

[3] M. Scholz, B. Claus (1999). Analysis and Simulation of Nonwoven Textures. Z. angew. Math. Mech. 79: 237-240. doi:10.1002/zamm.19990791362

[4] M. Lehmann, J. Eisengräber-Pabst, J. Ohser and A. Moghiseh (2013). Characterization of the Formation of Filter Paper using the Bartlett Spectrum of the Fiber Structure. Image Analysis & Stereology 32:77-87

[5] M. Godehardt, A. Moghiseh, C. Oetjen, J. Ohser and K. Schladitz (2021). An unambiguous cloudiness index for nonwovens. arXiv preprint.

[6] M. Deng, C. Dodson (1994). Paper: an engineered stochastic structure. Tappi Press, Atlanta

[7] B. Norman, D. Wahren (1974). The measurement of mass distribution in paper sheets using a beta radiograph method. Svensk Papperstilding 11: 397-406

[8] C. Chan, G. Pang (2000). Fabric defect detection by Fourier analysis. IEEE Transactions on Industry Applications 36(5):1267-1276

[9] J. Weickert (1996). A model for the cloudiness of fabrics. In: Progress in Industrial Mathematics at ECMI 94. Wiesbaden: Vieweg and Teubner, 258-265, doi: 10.1007/978-3-322-82967-2_31

[10] T. Penner, J. Meyer and A. Dittler (2021). Characterization of mesoscale inhomogeneity in nonwovens and its relevance in filtration of fine mists. J. Aerosol Sci. 151: 105674, doi: 10.1016/j.jaerosci.2020.105674.

[11] H.-C. Lien, C.-H. Liu (2006). A method of inspecting nonwoven basis weight using the exponential law of absorption and image processing. Textile Res. J. 76: 547-558

[12] A. Gourrier, W. Wagermaier, M. Burghammer, D. Lammie, H. S. Gupta, P. Fratzl, C. Riekel, T. J. Wess and O. Paris (2007). Scanning X-ray imaging with small-angle scattering contrast. J. Appl. Cryst. 40: 78-82, doi:10.1107/S0021889807006693